\documentclass[journal]{IEEEtran}
\usepackage{cite}
\usepackage{amsmath,amssymb,amsfonts}
\usepackage{graphicx}
\usepackage{textcomp}
\usepackage{xcolor}
\usepackage{caption}
\usepackage{subcaption}
\usepackage{multirow}
\usepackage{amssymb}
\usepackage{amsmath}
\usepackage{url}
\usepackage{soul}
\usepackage{color}
\usepackage{booktabs}
\usepackage{enumitem}
\usepackage{algorithm}
\usepackage[noend]{algpseudocode}

\definecolor{yellow}{rgb}{1.0, 0.65, 0.0}

\title{BiERU: Bidirectional Emotional Recurrent Unit \\for Conversational Sentiment Analysis}

\author{Wei Li,
        Wei Shao,
        Shaoxiong~Ji,
        and~Erik~Cambria,~\IEEEmembership{Senior Member,~IEEE}
        
\thanks{Wei Li, Nanyang Technological University, Singapore}
\thanks{Wei Shao, City University of Hong Kong, Hong Kong}
\thanks{Shaoxiong Ji, Aalto University, Finland}

\thanks{Corresponding Author: Erik Cambria (cambria@ntu.edu.sg), Nanyang Technological University, Singapore}
}

\date{}

\begin{document}
\maketitle
\begin{abstract}
Sentiment analysis in conversations has gained increasing attention in recent years for the growing amount of applications it can serve, e.g., sentiment analysis, recommender systems, and human-robot interaction. The main difference between conversational sentiment analysis and single sentence sentiment analysis is the existence of context information which may influence the sentiment of an utterance in a dialogue. How to effectively encode contextual information in dialogues, however, remains a challenge. Existing approaches employ complicated deep learning structures to distinguish different parties in a conversation and then model the context information. In this paper, we propose a fast, compact and parameter-efficient party-ignorant framework named bidirectional emotional recurrent unit for conversational sentiment analysis. In our system, a generalized neural tensor block followed by a two-channel classifier is designed to perform context compositionality and sentiment classification, respectively. Extensive experiments on three standard datasets demonstrate that our model outperforms the state of the art in most cases.
\end{abstract}


\section{Introduction}
\label{sec:intro}
\IEEEPARstart{S}{entiment} analysis and emotion recognition are of vital importance in dialogue systems and have recently gained increasing attention~\cite{maasur}. They can be applied to a lot of scenarios such as mining the opinions of speakers in conversations and improving the feedback of robot agents. Moreover, sentiment analysis in live conversations can be used in generating talks with certain sentiments to improve human-machine interaction. Existing approaches to conversational sentiment analysis can be divided into party-dependent approaches, like DialogueRNN~\cite{majumder2019dialoguernn}, and party-ignorant approaches, such as AGHMN~\cite{jiao2019real}. Party-dependent methods distinguish different parties in a conversation while party-ignorant methods do not.  Both party-dependent and party-ignorant models are not limited to dyadic conversations. Nevertheless, party-ignorant models can be easily applied to multi-party scenarios without any adjustment. In this paper, we propose a fast, compact and parameter-efficient party-ignorant framework based on the emotional recurrent unit (ERU), a recurrent neural network that contains a generalized neural tensor block (GNTB) and a two-channel feature extractor (TFE) to tackle conversational sentiment analysis. 
Nevertheless, party-ignorant models can be easily applied to multi-party scenarios without any adjustment. In this paper, we propose a fast, compact and parameter-efficient party-ignorant framework based on emotional recurrent unit (ERU), a recurrent neural network that contains a generalized neural tensor block (GNTB) and a two-channel feature extractor (TFE) to tackle conversational sentiment analysis.

\begin{figure}[t]
\centering\includegraphics[width=\linewidth]{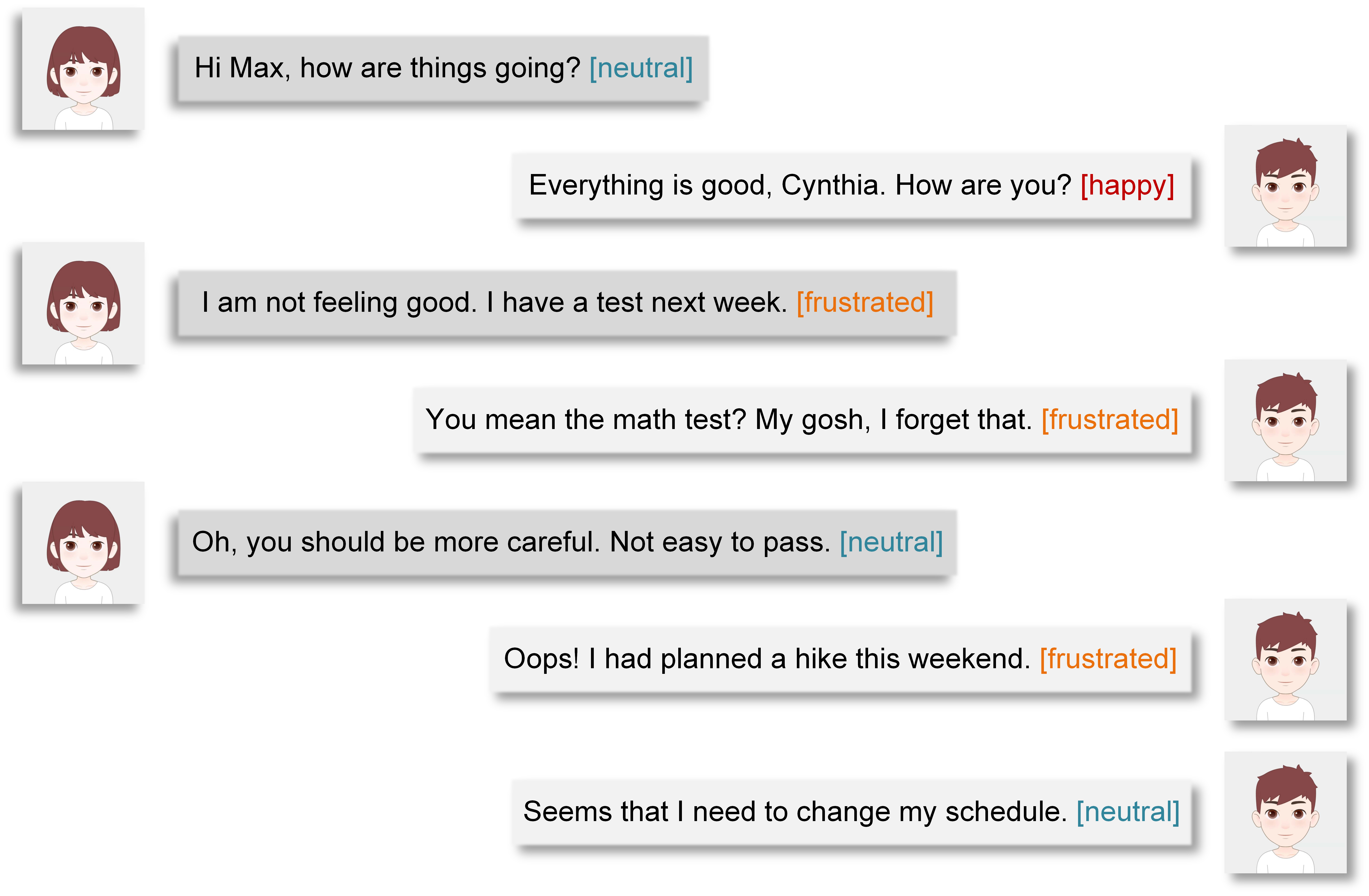}
\caption{Illustration of dialogue system and the interaction between talkers.}
\label{fig:DialogueSystem}
\end{figure}

Context information is the main difference between dialogue sentiment analysis and single sentence sentiment analysis tasks. It sometimes enhances, weakens, or reverses the raw sentiment of an utterance (Fig.~\ref{fig:DialogueSystem}). There are three main steps for sentiment analysis in a conversation: obtaining the context information, capturing the influence of the context information for an utterance, and extracting emotional features for classification. Existing dialogue sentiment analysis methods like c-LSTM~\cite{poria2017context}, CMN~\cite{hazarika2018conversational}, DialogueRNN~\cite{majumder2019dialoguernn}, and DialogueGCN~\cite{ghosal2019dialoguegcn} make use of complicated deep neural network structures to capture context information and describe the influence of context information for an utterance.

We redefine the formulation of conversational sentiment analysis and provide a compact structure to better encode the context information, capture the influence of context information for an utterance, and extract features for sentiment classification. According to Mitchell and Lapata~\cite{mitchell2010composition}, the meaning of a complete sentence must be explained in terms of the meanings of its subsentential parts, including those of its singular elements. Compositionality allows language to construct complicated meanings from its simpler terms. This property is often expressed in a manner of principle: the meaning of a whole is a function of the meaning of the components~\cite{partee1995lexical}. For conversation, the context of an utterance is composed of its historical utterances information. Similarly, context is a function of the meaning of its historical utterances. Therefore, inspired by the composition function in~\cite{partee1995lexical}, we design GNTB to perform context compositionality in conversation, which obtains context information and incorporates the context into utterance representation simultaneously, then employ TFE to extract emotional features. In this case, we convert the previous three-step task into a two-step task. Meanwhile, the compact structure reduces the computational cost. To the best of our knowledge, our proposed model is the first to perform context compositionality in conversational sentiment analysis.

The GNTB takes the context and current utterance as inputs and is capable of modeling conversations with arbitrary turns. It outputs a new representation of current utterance with context information incorporated (named `contextual utterance vector' in this paper). Then, the contextual utterance vector is further fed into TFE to extract emotional features. Here, we employ a simple two-channel model for emotion feature extraction.

The long short-term memory (LSTM) unit~\cite{hochreiter1997long} and one-dimensional convolutional neural network (CNN)~\cite{kim2014convolutional} are utilized for extracting features from the contextual utterance vector. Extensive experiments on three standard datasets demonstrate that our model outperforms state-of-the-art methods with fewer parameters. To summarize, the main contributions of this paper are as follows:

\begin{itemize}
    \item We propose a fast, compact and parameter-efficient party-ignorant framework based on ERU.
    \item We design GNTB which is suitable for different structures, to perform context compositionality.
    \item Experiments on three standard benchmarks indicate that our model outperforms the state of the art with fewer parameters.
\end{itemize}

The remainder of the paper is organized as follows: related work is introduced in Section~\ref{sec:related}; the mechanism of our model is explained in Section~\ref{sec:method}; results of the experiments are discussed in Section~\ref{sec:experiments}; finally, concluding remarks are provided in Section~\ref{sec:conclusion}.

\section{Related~Work}
\label{sec:related}
Sentiment analysis is one of the key natural language processing tasks that has drawn great attention from the research community in the last decade~\cite{cambig}. Besides the basic task of binary polarity classification~\cite{zhu2020sentivec}, sentiment analysis research has been carried out in many other related topics such as multimodal sentiment analysis~\cite{zadmul,ragima}, multilingual sentiment analysis~\cite{esucro}, aspect-based sentiment analysis~\cite{weiasp}, domain adaptation~\cite{banlex,xuuins}, rumors and fake news detection~\cite{akhnoo,reisup}, gender-specific sentiment analysis~\cite{mihwha,buktyp}, and multitask learning~\cite{yanseg}, including also applications of sentiment analysis in domains like healthcare~\cite{mahdet,qurmul}, political forecasting~\cite{ebrcha}, tourism~\cite{valsen}, customer relationship management~\cite{biicro}, stance classification~\cite{duucom}, and dialogue systems~\cite{wellea,schint}. Recently, some sophisticated deep learning techniques like capsule networks~\cite{zhang2020knowledge}, deep belief networks~\cite{liu2021speech}, and hybrid AI~\cite{camnt6} are applied to the research of sentiment analysis and aspect-based sentiment analysis.

Sentiment analysis in dialogues, in particular, has become a new trend recently. Poria et al.~\cite{poria2017context} proposed context-dependent LSTM networks to capture contextual information for identifying sentiment over video sequences, and Ragheb et al.~\cite{ragheb2019attention} utilized self-attention to prioritize important utterances.
Memory networks~\cite{sukhbaatar2015end}, which introduce an external memory module, was applied to modeling historical utterances in conversations. For example, CMN~\cite{hazarika2018conversational} modeled dialogue histories into memory cells, ICON~\cite{hazarika2018icon} proposed global memories for bridging self- and inter-speaker emotional influences, and AGHMN~\cite{jiao2019real} proposed hierarchical memory network as utterance reader.
{Ghosal et al.~\cite{ghosal2020cosmic} incorporated commonsense knowledge to enhance emotion recognition.} 
Recent advances in deep learning were also introduced to conversational sentiment analysis like attentive RNN~\cite{majumder2019dialoguernn}, adversarial training~\cite{wang2019capturing}, and graph convolutional networks~\cite{ghosal2019dialoguegcn}.
Another emerging direction is to incorporate Transformer-based contextual embedding.
Zhong et al.~\cite{zhong2019knowledge} leveraged commonsense knowledge from external knowledge bases to enrich transformer encoder. 
Qin et al.~\cite{qin2020dcr} built a co-interactive relation network to model feature interaction from bidirectional encoder representations from transformers (BERT) for joint dialogue act recognition and sentiment analysis.

Neural Tensor Networks (NTN)~\cite{socher2013reasoning} first proposed for reasoning over relational data are also related to our work. Socher et al.~\cite{socher2013recursive} further extended NTN to capture semantic compositionality for sentiment analysis. The authors proposed a tensor-based composition function to learn sentence representation recursively, which solves the issue when words function as operators that change the meaning of another word. 

\section{Method}
\label{sec:method}
\subsection{Problem Definition}

Given a multiple turns conversation $C$, the task is to predict the sentiment labels or sentiment intensities of the constituent utterances $U_1, U_2, ..., U_N$. Taking the interactive emotional database IEMOCAP~\cite{busso2008iemocap} as an example, emotion labels include frustrated, excited, angry, neutral, sad, and happy.

In general, the task is formulated as a multi-class classification problem over sequential utterances while in some scenarios, it is regarded as a regression problem given continuous sentiment intensity. In this paper, utterances are pre-processed and represented as $u_t$ using feature extractors described below.

\subsection{Textual Feature Extraction}
\label{sec:feature}
Following the tradition of DialogueRNN~\cite{majumder2019dialoguernn}, utterances are first embedded into vector space and then fed into CNN~\cite{kim2014convolutional} for feature extraction. N-gram features are obtained from each utterance by applying three different convolution filters of sizes 3, 4, and 5, respectively. Each filter has 50 features-maps. Majumder et al.~\cite{majumder2019dialoguernn} then uses max-pooling followed by rectified linear unit (ReLU) activation~\cite{nair2010rectified} to process the outputs of the convolution operation.

These activation values are concatenated and fed to a 100 dimensional fully connected layer whose outputs serve as the textual utterance representation. This CNN-based feature extraction network is trained at utterance level supervised by the sentiment labels.

\subsection{Our Model}

\begin{figure*}[tp]
\centering\includegraphics[width=0.95\textwidth]{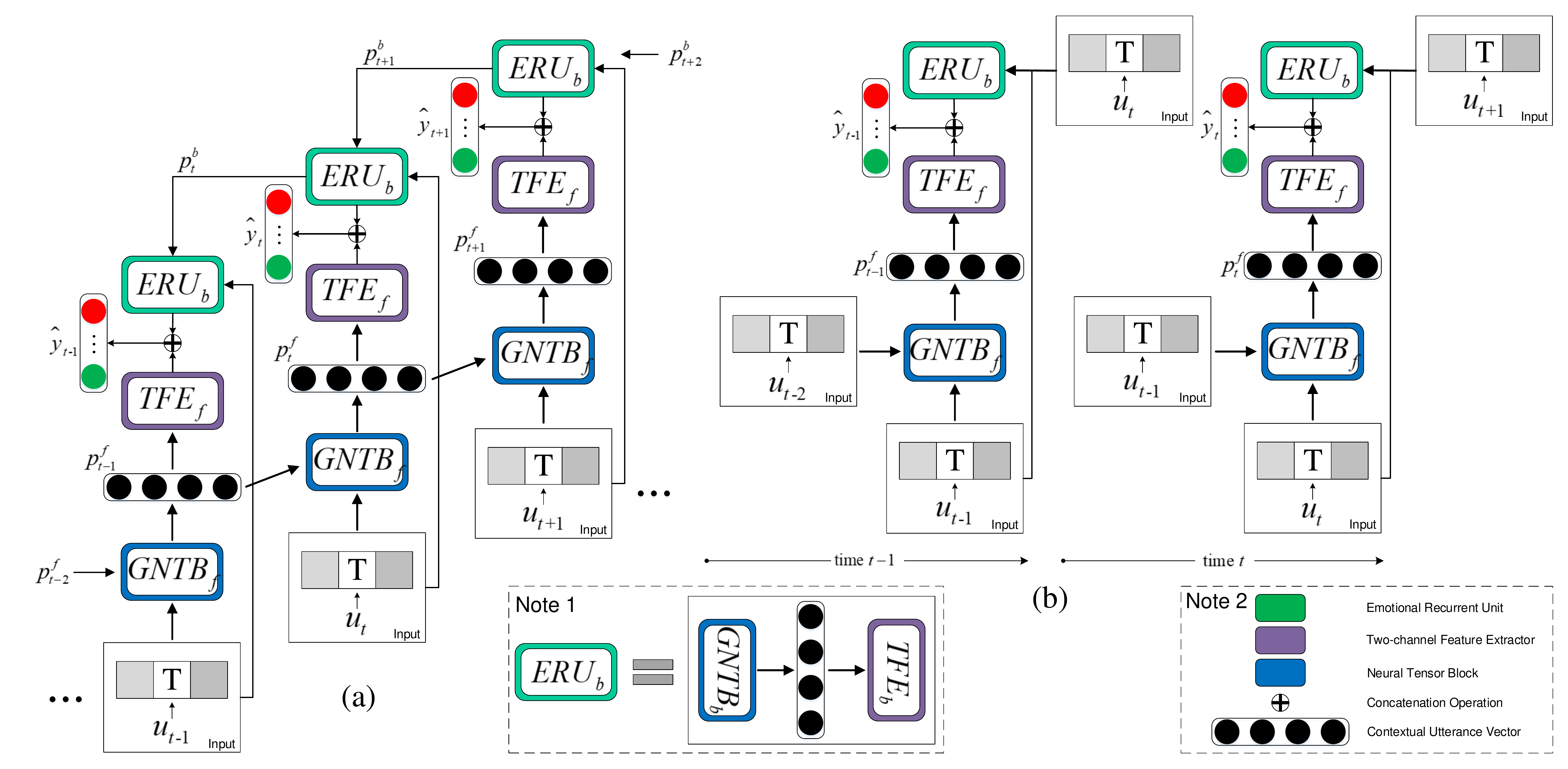}
\caption{(a) Architecture of BiERU with global context. (b) Architecture of BiERU with local context. Here $p^f_t$, $TFE_f$, and $GNTB_f$ are forward contextual utterance vector, TFE, and GNTB, respectively. $p^b_t$ and $ERU_b$ stand for backward contextual utterance vector and ERU, respectively. $\hat{y_t}$ is the predicted possibility vector of sentiment labels. T refers to textual modality in this paper. In our model, we only focus on textual modality. The detailed structures of GNTB and TFE are shown in Fig.~\ref{fig:ntb_and_tfe}.}
\label{fig:model}
\end{figure*}
Our ERU is illustrated in Note 1 of Fig.~\ref{fig:model}, which consists of two components GNTB and TFE. As mentioned in the introduction, there are three main steps for conversational sentiment analysis, namely obtaining the context representation, incorporating the influence of the context information into an utterance, and extracting emotional features for classification. In this paper, the ERU is employed in a bidirectional manner (BiERU) to conduct the above sentiment analysis task, reducing some expensive computations and converting the previous three-step task into a two-step task as shown in Fig.~\ref{fig:model}.

Similar to bidirectional LSTM (BiLSTM)~\cite{graves2005framewise}, two ERUs are utilized for forward and backward passing the input utterances. Outputs from the forward and backward ERUs are concatenated for sentiment classification or regression. More concretely, the GNTB is applied to encoding the context information and incorporating it into an utterance simultaneously; while TFE takes the output of GNTB as input and is used to obtain emotional features for classification or regression.

\subsubsection{Generalized Neural Tensor Block}
The utterance vector $u_t \in R^{d}$ with the context information incorporated is named as contextual utterance vector $p_t \in R^{d} $ in this paper, where $d$ is the dimension of $u_t$ and $p_t$. At time $t$, GNTB (Fig.~\ref{fig:ntb_and_tfe}: (a)) takes $u_t$ and $p_{t-1}$ as inputs and then outputs $p_t$, a contextual utterance vector. In this process, GNTB first extracts the context information from $p_{t-1}$; it then incorporates the context information into $u_t$; finally, contextual utterance vector $p_t$ is obtained. The first step is to capture the context information and the second step is to integrate the context information into current utterance. The combination of these two steps is regarded as context compositionality in this paper. To the best of our knowledge, this is the first work to perform context compositionality in conversational sentiment analysis. GNTB is the core part that achieves the context compositionality. The formulation of GNTB is described below:
\begin{equation}
    \label{E1}
    p_t = f(m_t^T T^{[1:k]} m_t + W m_t)
\end{equation}
\begin{equation}
    \label{E2}
     m_t = p_{t-1} \oplus u_t
\end{equation}
where $m_t \in R^{2d}$ is the concatenation of $p_{t-1}$ and $u_t$; $f$ is an activation function, such as $tanh$ and $sigmoid$; the tensor $T^{[1:k]} \in R^{2d \times 2d \times k}$ and the matrix $W \in R^{k\times 2d}$ are the parameters used to calculate $p_t$. Each slice $T^{[i]} \in R^{2d \times 2d}$ can be interpreted as capturing a specific type of context compositionality. Each slice $W^{[i]} \in R^{1 \times 2d}$ maps contextual utterance vector $p_{t}$ and utterance vector $u_{t}$ into the context compositionality space. Here we have $k$ different context compositionality types, which constitutes $k$-dimensional context compositionality space. The main advantage over the previous NTN~\cite{socher2013reasoning}, which is a special case of the GNTB when $k$ is set to $d$, is that GNTB is suitable for different structures rather than only the recursive structure and the space complexity of GNTB is $O(kd^2)$ compared with $O(d^3)$ in NTN. In order to further reduce the number of parameters, we employ the following low-rank matrix approximation for each slice $T^{[i]}$:
\begin{equation}
    \label{E3}
     T^{[i]} = UV + diag(e)
\end{equation}
where $U \in R^{2d \times r}$, $V \in R^{r \times 2d}$, $e \in R^{2d}$ and $r \ll d$.

\subsubsection{Two-channel Feature Extractor}

\begin{figure}[!ht]
\centering\includegraphics[width=0.4\textwidth]{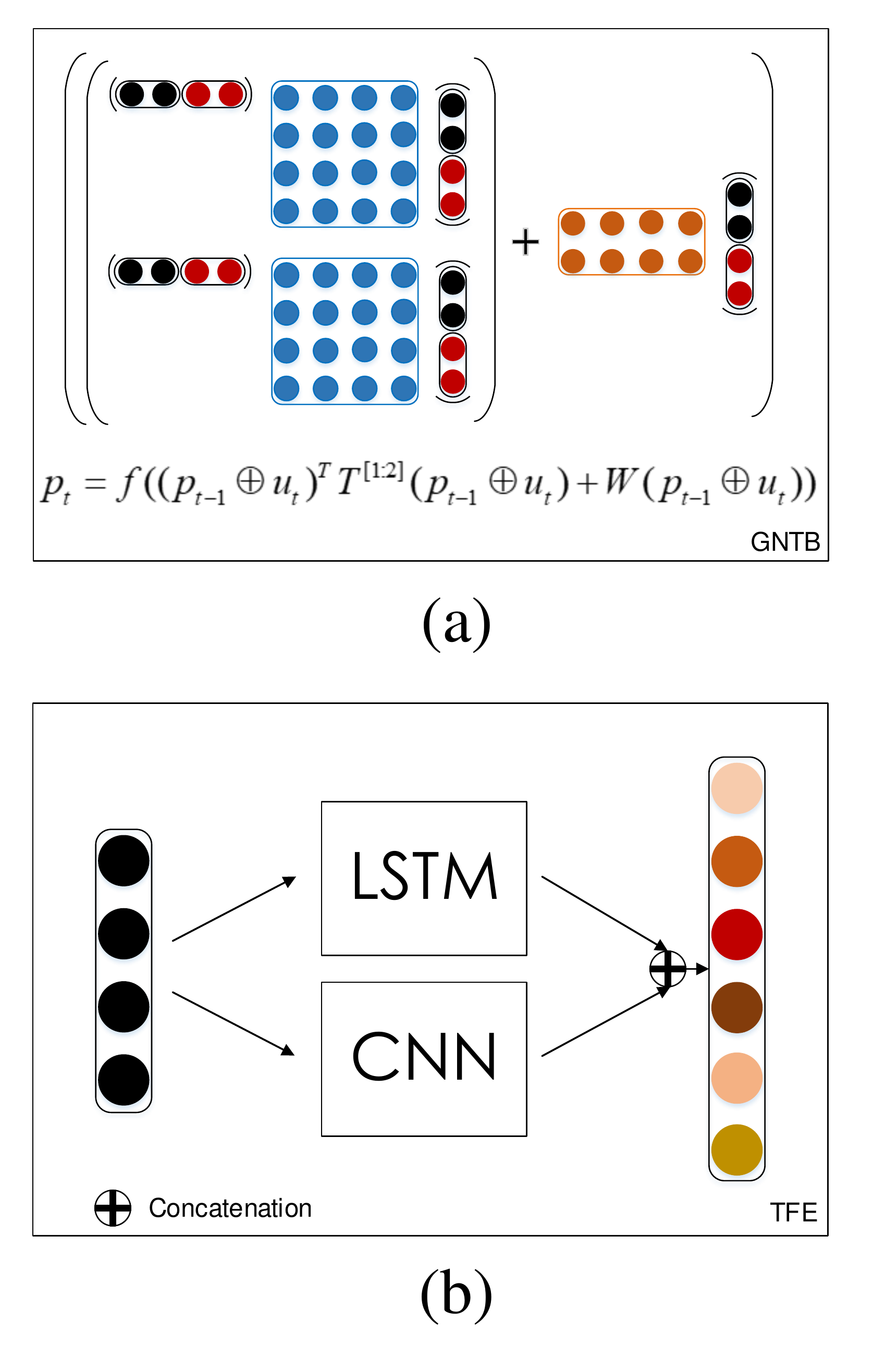}
\caption{(a) GNTB when $u_t \in R^{2}$. (b) TFE. The input of LSTM and CNN is context utterance vector $p_t$, and output is emotion features $e_t$.}
\label{fig:ntb_and_tfe}
\end{figure}

We utilize TFE to refine the emotion features from contextual vector $p_t$. As shown in Fig.~\ref{fig:ntb_and_tfe}: (b), the TFE is a two-channel model, including an  LSTM cell~\cite{hochreiter1997long} branch and a one-dimensional CNN~\cite{kim2014convolutional} branch. The two branches receive the same contextual utterance vector $p_t$ and produce outputs that may contain complementary information~\cite{li2020user}.

At time $t$, the LSTM cell takes hidden state $h_{t-1}$, cell state $c_{t-1}$ and the contextual utterance vector $p_t$ as inputs, where $h_{t-1}$ and $c_{t-1}$ are obtained from the last time step $t-1$. The outputs of the LSTM cell are updated hidden state $h_t$ and cell state $c_t$. The hidden state $h_t$ is regarded as the emotion feature vector. The CNN receives $p_t$ as input and outputs the emotion feature vector $l_t$. Finally, the outputs of LSTM cell branch $h_t$ and CNN branch $l_t$ are concatenated into an emotion feature vector $e_t$ which is also the output of ERU. The formulas of TFE are as follows:
\begin{equation}
    h_t, c_t = \mathop{LSTMCell}(p_t, (h_{t-1}, c_{t-1})
\end{equation}
\begin{equation}
    l_t = \mathop{CNN}(p_t)
\end{equation}
\begin{equation}
    e_t = h_t \oplus l_t
\end{equation}
\subsubsection{Sentiment Classification \& Regression}
Taking emotion feature $e_t$ as input, we use a linear neural network $W_c \in R^{D_e \times n\_class}$ followed by a softmax layer to predict the sentiment labels, where $n\_class$ is the number of sentiment labels.

Then, we obtain the probability distribution $S_t$ of the sentiment labels. Finally, we take the most possible sentiment class as the sentiment label of the utterance $u_t$:
\begin{equation}
    S_t = \mathop{Softmax}(W_c^\mathrm{T} e_t)
\end{equation}
\begin{equation}
    \hat{y_t} = \arg \max \limits_{i} (S_t[i])
\end{equation}
For sentiment regression task, we use a linear neural network $W_r \in R^{D_e \times 1}$ to predict the sentiment intensity. Then, we obtain the predicted sentiment intensity $q_t$:
\begin{equation}
    q_t = W_r^\mathrm{T} e_t
\end{equation}
where $W_s \in R^{D_e \times n\_class}$, $e_t \in R^{D_e}$, $S_t \in R^{n\_class}$, $q_t$ is a scalar and $\hat{y_t}$ is the predicted sentiment label for utterance $u_t$.

\subsubsection{Training}
For the classification task, we choose cross-entropy as the measure of loss and use L2-regularization to relieve overfitting. The loss function is:
\begin{equation}{}
L=-\frac{1}{\sum_{s=1}^{N} c(s)} \sum_{i=1}^{N} \sum_{j=1}^{c(i)} \log S_{i, j}\left[y_{i, j}\right]+\lambda\|\theta\|_{2}
\end{equation}

For the regression task, we choose the mean square error (MSE) to measure loss, and L2-regularization to relieve overfitting. The loss function is:
\begin{equation}{}
L=\frac{1}{\sum_{s=1}^{N} c(s)} \sum_{i=1}^{N} \sum_{j=1}^{c(i)}\left( q_{i, j} - z_{i, j}\right)^{2}+\lambda\|\theta\|_{2}
\end{equation}
where N is the number of samples/conversations, $S_{i,j}$ is the probability distribution of sentiment labels for utterance $j$ of conversation $i$, $y_{i,j}$ is the expected class label of utterance $j$ of conversation $i$, $q_{i,j}$ is the predicted sentiment intensity of utterance $j$ of conversation $i$, $z_{i,j}$ is the expected sentiment intensity of utterance $j$ of conversation $i$, c(i) is the number of utterances in sample $i$, $\lambda$ is the L2-regularization weight, and $\theta$ is the set of trainable parameters. We employ stochastic gradient descent based Adam~\cite{kingma2014adam} optimizer to train our network.

\subsection{Bidirectional Emotion Recurrent Unit Variants}
Our model has two different forms according to the source of context information, namely bidirectional emotion recurrent unit with global context (BiERU-gc) and bidirectional emotion recurrent unit with local context (BiERU-lc).

\subsubsection{BiERU-gc}
According to equation (\ref{E1}), GNTB extracts the context information from $p_{t-1}$, integrates the context information into $u_t$, and thus obtains the contextual utterance vector $p_t$. Based on the definition of contextual utterance vector, $p_{t-1}$ is the utterance vector that contains information of $u_{t-1}$ and $p_{t-2}$. In this case, the contextual utterance vector $p_t$ holds the context information from all the preceding utterances $u_1, u_2,\cdots, u_{t-1}$ in a recurrent manner. 
{Bidirectional neural networks have empirically gained improved performance than its counterpart with only forward propagation~\cite{schuster1997bidirectional}. As shown in Fig.~\ref{fig:model} : (a), we utilize the bidirectional setting to capture context information from surrounding utterances.}
The BiERU in Fig.~\ref{fig:model} :(a) is named as BiERU-gc.
\subsubsection{BiERU-lc}
Following equation (\ref{E1}), GNTB extracts the context information from the contextual utterance vector $p_{t-1}$, and $p_{t-1}$ contains the context information of all the preceding utterances $u_1, u_2, \cdots, u_{t-2}$ as mentioned above. If replacing $p_{t-1}$ with $u_{t-1}$ in equation (\ref{E1}) and (\ref{E2}), $p_t$ contains the information of $u_{t-1}$ and $u_t$. In other words, $u_{t-1}$ is not only an utterance vector, but also works as the context of $u_t$. As shown in Fig.~\ref{fig:model} : (b), bidirectional ERU makes $p_t$ obtain the future information $u_{t+1}$. In this case, GNTB extracts the context information from $u_{t-1}$ and $u_{t+1}$, which are the adjacent utterances of $u_t$. We name this model as BiERU-lc.

\section{Experiments}
\label{sec:experiments}
In this section, we conduct a series of comparative experiments to evaluate the performance of our proposed model (Codes are available on our GitHub\footnote{https://github.com/Maxwe11y/BiERU.}) and perform a thorough analysis.

\subsection{Datasets}
We use three datasets for experiments, i.e., AVEC~\cite{schuller2012avec}, IEMOCAP~\cite{busso2008iemocap} and MELD~\cite{poria2018meld}, which are also used by some representative models such as DialogueRNN~\cite{majumder2019dialoguernn} and DialogueGCN~\cite{ghosal2019dialoguegcn}. We conduct the standard data partition rate (details in Table~\ref{tab:data}).
\begin{table}[!htb]
    \centering
    \begin{tabular}{|c|c|c|c|}
        \hline
        DATASET & Partition & \ Utterance Count & \ Dialogue Count \\
        \hline
        \hline
        \multirow{2}{*}{IEMOCAP} & train + val & 5810 & 120 \\
        \cline{2-4}
        & test & 1623 & 31 \\
        \hline
        \multirow{2}{*}{AVEC} & train + val & 4368 & 63 \\
        \cline{2-4}
        & test & 1430 & 32 \\
        \hline
        \multirow{2}{*}{MELD} & train + val & 11098 & 1153 \\
        \cline{2-4}
        & test & 2610 & 280 \\
        \hline
    \end{tabular}
    \caption{Statistical information and data partition of datasets used in this paper.}
    \label{tab:data}
\end{table}{}

Originally, these three datasets are multimodal datasets. Here, we focus on the task of textual conversational sentiment analysis and only use the textual modality to conduct our experiments.

\paragraph{IEMOCAP}
The IEMOCAP~\cite{busso2008iemocap} is a dataset of two-way conversations involved with ten distinct participators. It is recorded as videos where every video clip contains a single dyadic dialogue, and each dialogue is further segmented into utterances. Each utterance is labeled as one sentiment label from six sentiment labels, i.e., happy, sad, neutral, angry, excited and frustrated~\cite{susanto2020hourglass}. The dataset includes three modalities: audio, textual and visual. Here, we only use textual modality data in experiments.

\paragraph{AVEC}
The AVEC dataset~\cite{schuller2012avec} is a modified version of the SEMAINE database~\cite{mckeown2011semaine} that contains interactions between human speakers  and robots. Unlike IEMOCAP, each utterance in the AVEC dataset is given an annotation every 0.2 second with one of four real valued attributes, i.e., valence ($\left[-1,1\right]$), arousal ($\left[-1,1\right]$), expectancy ($\left[-1,1\right]$), and power ($\left[0,\infty\right]$). Our experiments use the processed utterance-level annotation~\cite{majumder2019dialoguernn}, and treat four affective attributes as four subsets for evaluation.

\paragraph{MELD}
The MELD~\cite{poria2018meld} is a multimodal and multiparty sentiment analysis /classification database. It contains textual, acoustic, and visual information for more than 13000 utterances from the Friends TV series. The sentiment label of each utterance in a dialogue lies within one of the following seven sentiment classes: fear, neutral, anger, surprise, sadness, joy and disgust.
\subsection{Baselines~and~Settings}
To evaluate the performance of our model, we choose the following models as strong baselines including the state-of-the-art methods.


\paragraph{c-LSTM~\cite{poria2017context}}
The c-LSTM uses bidirectional LSTM~\cite{hochreiter1997long} to learn contextual representation from the surrounding utterances. When combined with the attention mechanism, it becomes the c-LSTM+Att.


\paragraph{CMN~\cite{hazarika2018conversational}}
This model utilizes memory networks and two different GRUs~\cite{cho2014learning} for two speakers for representation learning of utterance context from dialogue history.

\paragraph{DialogueRNN~\cite{majumder2019dialoguernn}}
It distinguishes different parties in a conversation interactively, with three GRUs representing the speaker states, context, and emotion.
It has several variants including DialogueRNN+Att with attention mechanism and bidirectional BiDialgoueRNN.

\paragraph{DialogueGCN~\cite{ghosal2019dialoguegcn}}
This model employs graph neural networks based approach through which context propagation issue can be addressed, to detect sentiment in conversations.
\paragraph{AGHMN~\cite{jiao2019real}}
It utilizes hierarchical memory networks with BiGRUs for utterance reader and fusion, and attention mechanism for memory summarizing.

\paragraph{Settings}
All the experiments are performed using CNN extracted features as described in the Method section. For a fair comparison with the state-of-the-art DialogueRNN model, we use their utterance representation directly\footnote{Extracted features of two datasets are available at \url{https://github.com/senticnet/conv-emotion}.}.

To alleviate over-fitting, we employ
Dropout~\cite{srivastava2014dropout} over the outputs of GNTB and TFE. For the nonlinear activation function, we choose the sigmoid function for sentiment classification and the ReLU function for sentiment regression. Our model is optimized by an Adam optimizer~\cite{kingma2014adam}. Hyper-parameters are tuned manually. Batch size is set as 1. We set the rank for all the experiments to $r=10$. Our model is implemented using PyTorch~\cite{paszke2019pytorch}. In Table \ref{tab:param}, we display the hyper-parameters of our BiERU-lc model on the three standard datasets.

\begin{table}[!htb]
    \centering
    \begin{tabular}{|c|c|c|c|}
        \hline
        \multirow{2}{*}{DATASET} & Dropout & \ Learning & \ Regularization \\
        & Rate & Rate & Weight\\
        \hline
        \hline
        IEMOCAP & 0.8 & 0.0001 & 0.001 \\
        \hline
        AVEC.VALENCE & 0.5 & 0.0001 & 0.0002 \\
        AVEC.AROUSAL & 0.8 & 0.0001 & 0.0002  \\
        AVEC.EXPECTANCY & 0.5 & 0.00005 & 0.0005  \\
        AVEC.POWER & 0.8 & 0.0001 & 0.0001  \\
        \hline
        MELD & 0.7 & 0.0005 & 0.001 \\
        \hline
    \end{tabular}
    \caption{Hyper-parameters of our BiERU-lc model on different datasets.}
    \label{tab:param}
\end{table}{}

\linespread{1.25}
\begin{table*}[!ht]
    \centering
    \resizebox{1.0\textwidth}{20mm}{
    \begin{tabular}{|c||cc|cc|cc|cc|cc|cc|cc||c|}
    \hline
    \multirow{3}{*}{METHODS} & \multicolumn{14}{c||}{IEMOCAP} &
    \multicolumn{1}{c|}{MELD} \\
    \cline{2-16}
    & \multicolumn{2}{c|}{Happy} & \multicolumn{2}{c|}{Sad} & \multicolumn{2}{c|}{Neutral} & \multicolumn{2}{c|}{Angry} & \multicolumn{2}{c|}{Excited} & \multicolumn{2}{c|}{Frustrated} & \multicolumn{2}{c||}{Average} &
    \multicolumn{1}{c|}{Average} \\
    \cline{2-16}
	&	Acc. &	F1 &  Acc.	& F1 & Acc.	& F1 & Acc.	&	F1 &  Acc.	&	F1 & Acc.	&	F1 &  Acc.	&	F1  & Acc.	\\

\hline
c-LSTM & 30.56 & 35.63 & 56.73  & 62.90 & 57.55 & 53.00 & 59.41 & 59.24 & 52.84 & 58.85 & 65.88 & 59.41 & 56.32 & 56.19 & 57.5    \\
\hline
CMN & 25.00 & 30.38 & 55.92 & 62.41 & 52.86 & 52.39 & 61.76 & 59.83 & 55.52 & 60.25 & \textbf{71.13} & 60.69 & 56.56 & 56.13 & -    \\
\hline
DialogueRNN	&	25.69 & 33.18	&	75.10 & 78.80	&	58.59 & 59.21	&	64.71 & \textbf{65.28}	&	\textbf{80.27} & 71.86	&	61.15 & 58.91	&	63.40	&	62.75 &  56.1 \\
\hline
DialogueGCN	&	40.62  & \textbf{42.75}	&	\textbf{89.14} & \textbf{84.54}	&	61.92 & \textbf{63.54}	&	67.53 & 64.19	&	65.46 & 63.08	&	64.18 & \textbf{66.99}	&	65.25	&	64.18	& - \\
\hline
AGHMN	&	48.30 & \textbf{52.1}	&	68.30 & 73.3	&	61.60 & 58.4	&	57.50 & 61.9	&	\textbf{68.10} &  69.7	&	\textbf{67.10} & \textbf{62.3}	&	63.50	&	63.50  & 60.3 \\
\hline
BiERU-gc	&	\textbf{49.81} & 32.75	&	\textbf{81.26} & 82.37	&	\textbf{65.00} & \textbf{60.45}	&	\textbf{67.86} & \textbf{65.39}	&	63.14 & \textbf{73.29}	&	59.77 & 60.68	&	\textbf{65.35} & \textbf{64.24} & \textbf{60.7}    \\
\hline
BiERU-lc	& \textbf{55.44} & 31.56	&	80.19  & \textbf{84.13}	&	\textbf{64.73} & 59.66	&	\textbf{69.05} & 65.25	&	63.18 & \textbf{74.32}	&	61.06 & 61.54	&	\textbf{66.09}	&	\textbf{64.59}	& \textbf{60.9}\\
    \hline
    \end{tabular}}
    \linespread{1}
    \caption{Comparison with baselines on IEMOCAP and MELD datasets using textual modality. Average score of accuracy and f1-score are weighted. ``-'' represents no results reported in original paper.}
    \label{tab:results}
\end{table*}

\begin{table}[!ht]
    \centering
    \begin{tabular}{|c|c|c|c|c|}
    \hline
    \multirow{3}{*}{METHODS} & \multicolumn{4}{c|}{AVEC} \\
    \cline{2-5}
    & \multicolumn{1}{c|}{Valence} & \multicolumn{1}{c|}{Arousal} & \multicolumn{1}{c|}{Expectancy} & \multicolumn{1}{c|}{Power} \\
    \cline{2-5}
	&	r	&	r	&	r	&	r \\

\hline
c-LSTM & 0.16 & 0.25 & 0.24 & 0.10   \\
\hline
CMN &  0.23 & 0.29 &  0.26 & -0.02  \\
\hline
DialogueRNN	&	0.35	&	0.59	&	0.37	&	\textbf{0.37}   \\
\hline
BiERU-gc	& 0.30	&	0.63	&	0.36	&	0.36  \\
\hline
BiERU-lc	& \textbf{0.36}	&	\textbf{0.64}	&	\textbf{0.38}	&	\textbf{0.37} \\
    \hline
    \end{tabular}
    \linespread{1}
    \caption{Comparison with baselines on AVEC dataset using textual modality. $r$ stands for Pearson correlation coefficient.}
    \label{tab:result_avec}
\end{table}

\subsection{Results}
We compare our model with baselines on textual modality using three standard benchmarks. We run the experiment five times and report the average results. Overall, our model outperforms all the baseline methods including state-of-the-art models like DialogueRNN, DialogueGCN and AGHMN on these datasets, and markedly exceeds in some indicators as the results show in Table~\ref{tab:results}. 

For the IEMOCAP dataset as a classification problem, we use accuracy for each class, and weighted average of accuracy and f1-score for measuring the overall performance. As for the AVEC dataset, standard metrics for the regression task including Pearson correlation coefficient ($r$) are used for evaluation. We use weighted average of accuracy as the measure of performance on the MELD dataset.

\subsubsection{Comparison with the State of the Art}
We firstly compare our proposed BiERU with state-of-the-art methods DialogueGCN, DialogueRNN and AGHMN on IEMOCAP, AVEC and MELD, respectively.

\paragraph{IEMOCAP}
As shown in Table~\ref{tab:results}, our proposed BiERU-gc model exceeds the best model DialogueGCN by $0.10\%$ and $0.06\%$ in terms of weighted average accuracy and f1-score, respectively. And the BiERU-lc model pushes up state-of-the-art results by $0.84\%$ and $0.41\%$ for weighted average accuracy and f1-score, respectively. For all 14 indicators on the IEMOCAP dataset, our models outperform at 7 indicators and have more balanced performances over these six classes. In particular, the accuracy of ``happy'' of our proposed BiERU-lc is higher than the result of DialgoueGCN by $14.82\%$. In the DialogueGCN model, the authors employ a two-layer graph convolutional network to model the interactions between speakers within a sliding window. For dyadic conversations, there are 4 different relations and context information is scattered into each relation. In this case, however, the context information is incomplete and inadequate for each relation. Besides, window size is fixed, which makes it inflexible to different scenarios. Our proposed models, to some extent, is more capable of capturing adequate context information. To sum up, the experimental results indicate that BiERU models can effectively capture contextual information and extract rich emotion features to boost the overall performance and achieve relatively balanced results. 

\paragraph{AVEC}
Among these four attributes, our model outperforms DialogueRNN for ''valence'', ``arousal'' and ``expectancy'' attributes and obtains the same results on the ''power'' attribute. The pearson correlation coefficient $r$ of BiERU-gc is $0.04$ higher than its counterpart in terms of ``arousal'' (Table~\ref{tab:result_avec}). As for the BiERU-lc model, it is $0.05$ higher in $r$. For the attributes ``expectancy'' and ''valence'', the BiERU-lc model is $0.01$ higher in $r$. As for the attribute``power'', although our best model does not outperform the state-of-the-art method, it surpasses most of the other baseline methods including CMN and c-LSTM.
Overall, the BiERU-lc model works well on all the attributes, considering the benchmark performances are very high.  As mentioned in part~\ref{tab:data} of section~\ref{sec:experiments}, AVEC is composed of conversations between human speakers and robots. Robots are not good at identifying global information and tend to respond to adjacent queries from human speakers. This is one possible reason that our BiERU-lc model has better performances than baselines and BiERU-gc since it is skilled at capturing local context information.

\paragraph{MELD}
Three factors make it considerably harder to model sentiment analysis on MELD in comparison with IEMOCAP and AVEC datasets. First, the average number of turns in a MELD conversation is 10 while it is close to 50 on the IEMOCAP. Second, there are more than 5 speakers in most of the MELD conversations, which means most of the speakers only utter one or two utterances per conversation. What's worse, sentiment expressions rarely exist in MELD utterances and the average length of MELD utterances is much shorter than it is in IEMOCAP and AVEC datasets. For a party-dependent model like DialogueRNN, it is hard to model inter-dependency between speakers. We find that the performances of party-ignorant models such as c-LSTM and AGHMN are slightly better than party-dependent models on this dataset. Our BiERU models utilize GNTB to perform context compositionality and achieve the state-of-the-art average accuracy of $60.9\%$, outperforming AGHMN by $0.6\%$ and DialogueRNN by $4.8\%$.

\subsubsection{Comparison between BiERU-gc and BiERU-lc}
The proposed two variants take different context inputs. The BiERU-gc model takes the output of GNTB at the last time step and the current utterance as the input of GNTB at the current time step. And the BiERU-lc model uses the last utterance and current utterance as input of GNTB at the current time step. According to experimental results in Tables~\ref{tab:results} and~\ref{tab:result_avec}, the overall performance of BiERU-lc is better than BiERU-gc.

For IEMOCAP datasets, the BiERU-lc model surpasses the BiERU-gc model by $0.74\%$ and $0.45\%$ in terms of weighted average accuracy and f1-score, respectively. For the AVEC and MELD datasets, BiERU-lc also outperforms its counterpart. One possible explanation is that context information of a contextual utterance vector in BiERU-gc comes from all utterances in the current conversation. However, in BiERU-lc, the context information comes from neighborhood utterances. In this case, context information of BiERU-gc contains redundant information and thus has a negative impact on emotion feature extraction.

\begin{figure}[!ht]
    \centering
	\includegraphics[width=0.5\textwidth]{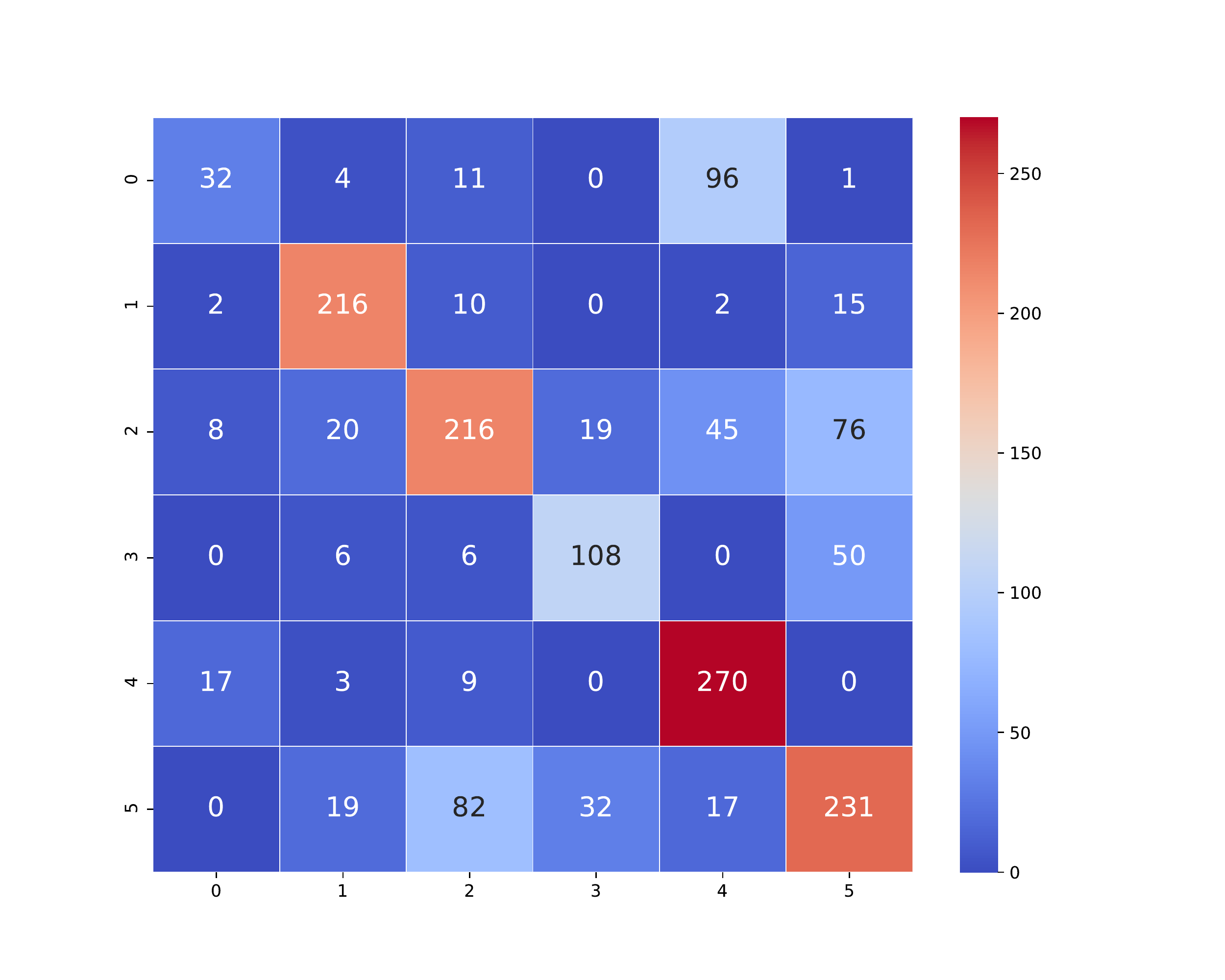}
	\linespread{1}
    \caption{Heat map of confusion matrix of BiERU-lc.}
\label{fig:heatmap}
\end{figure}{}

\subsection{Case~Study}
\begin{figure*}[!ht]
    \centering
	\includegraphics[width=1.00\textwidth]{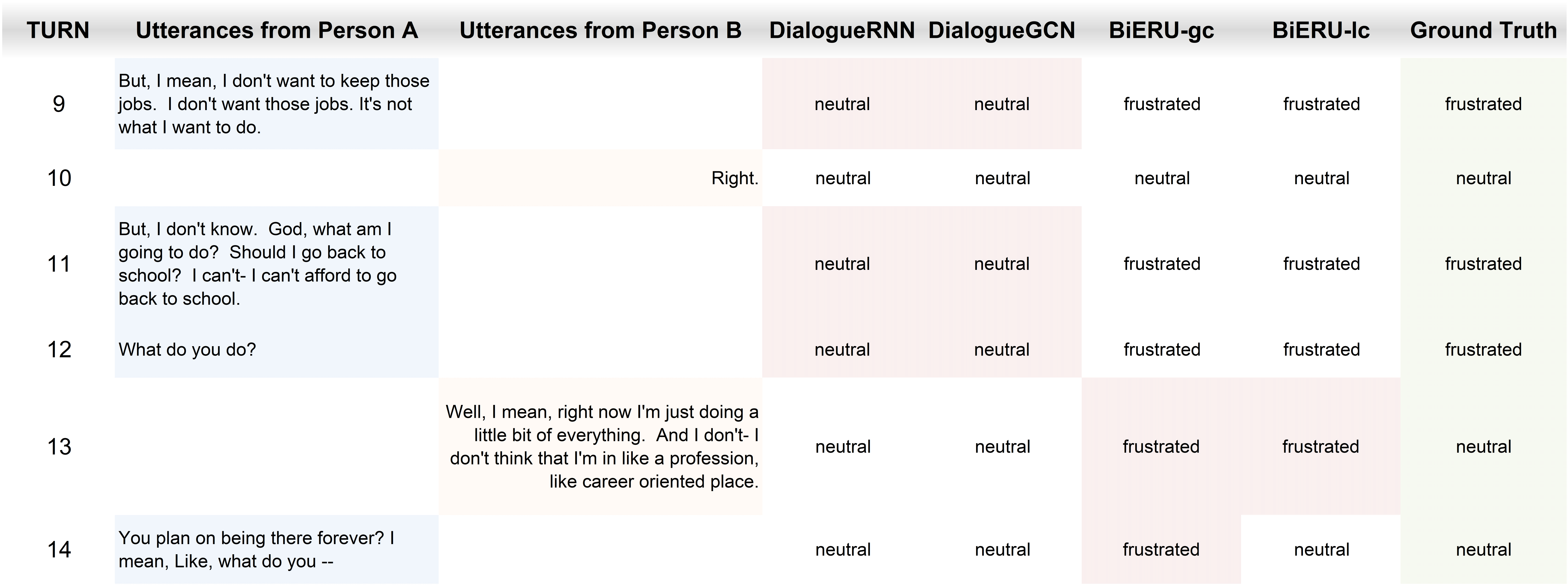}
	\linespread{1}
\caption{Illustration of a conversation snippet from IEMOCAP dataset.}
\label{fig:case_study}
\end{figure*}

Figure~\ref{fig:case_study} illustrates a conversation snippet classified by our BiERU-lc method. In this snippet, person A is initially in a frustrated state while person B acts as a listener in the beginning. Then, person A changes his/her focus and questions person B on his/her job state. Person B tries to use his/her own experience to help person A get rid of the frustrating state. This snippet reveals that the sentiment of a speaker is relatively steady and the interaction between speakers may change the sentiment of a speaker. Our BiERU-lc method shows good ability in capturing the speaker's sentiment (turns 9, 11, 12, 14) and the interaction between speakers (turn 10). The sentiment in turn 13 is very subtle. Turn 13 contains a little bit of frustration since he/she is not satisfied with his/her job state. However, considering that person B attempts to help person A, turn 13 is more likely to be in a neutral stand. Besides, we also display the prediction results of baselines including the state of the art in Fig.~\ref{fig:case_study}. On the one hand, both the DialogueRNN and DialogueGCN models cannot successfully model the interaction between the two speakers in this dialogue snippet. On the other hand, the two baselines are more likely to classify a few consecutive utterances into the same emotion label, which indicates that they are insensitive to the context information. In contrast, our BiERU-gc model gets better results and detects the emotion shifting from frustrated to neutral and from frustrated to neutral. However, global context may contain noise information that is not related to the current utterance, which weakens the proportion of related information and makes the BiERU-gc model less sensitive to sentiment shifting. In this case, the BiERU-lc model obtains the best results on this dialogue snippet since it is more sensitive to the context information and has a better context compositionality ability in general.

{}


\subsection{Visualization}
We use visualization to provide some insights into the proposed model. Firstly, we visualize the confusion matrix in the form of a heat map to describe the performance of our BiERU-lc model. The heat maps of BiERU-lc on the IEMOCAP dataset are shown in Fig.~\ref{fig:heatmap}.
Our model has a balanced performance over all the sentiment classes.

\begin{figure}[ht!]
\centering
\begin{subfigure}{0.45\textwidth}
    \includegraphics[width=\linewidth]{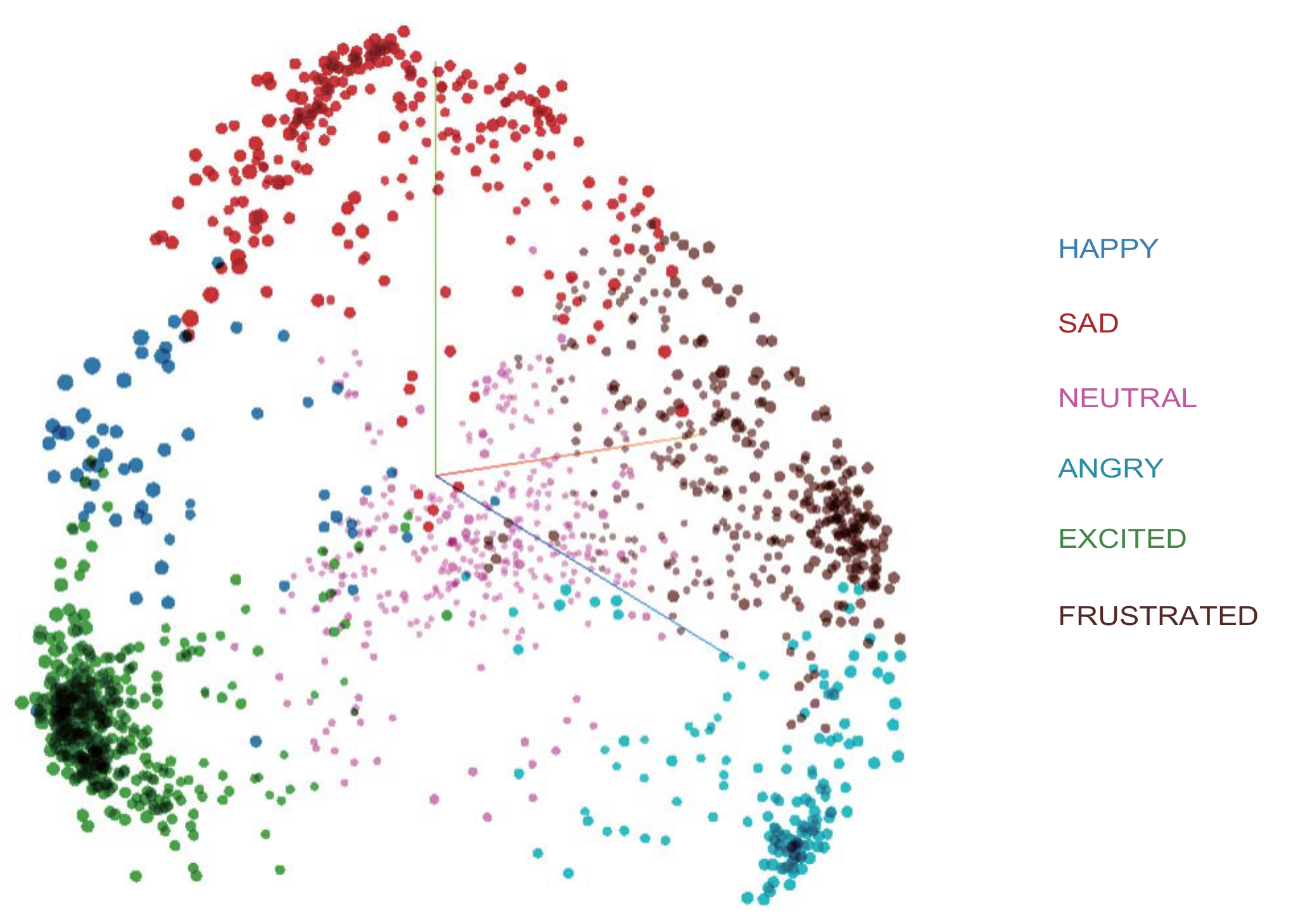}
    \subcaption{BiERU-lc}
    \label{fig:Dimensionality Reduction of BiERU}
\end{subfigure}
\begin{subfigure}{0.45\textwidth}
    \includegraphics[width=\linewidth]{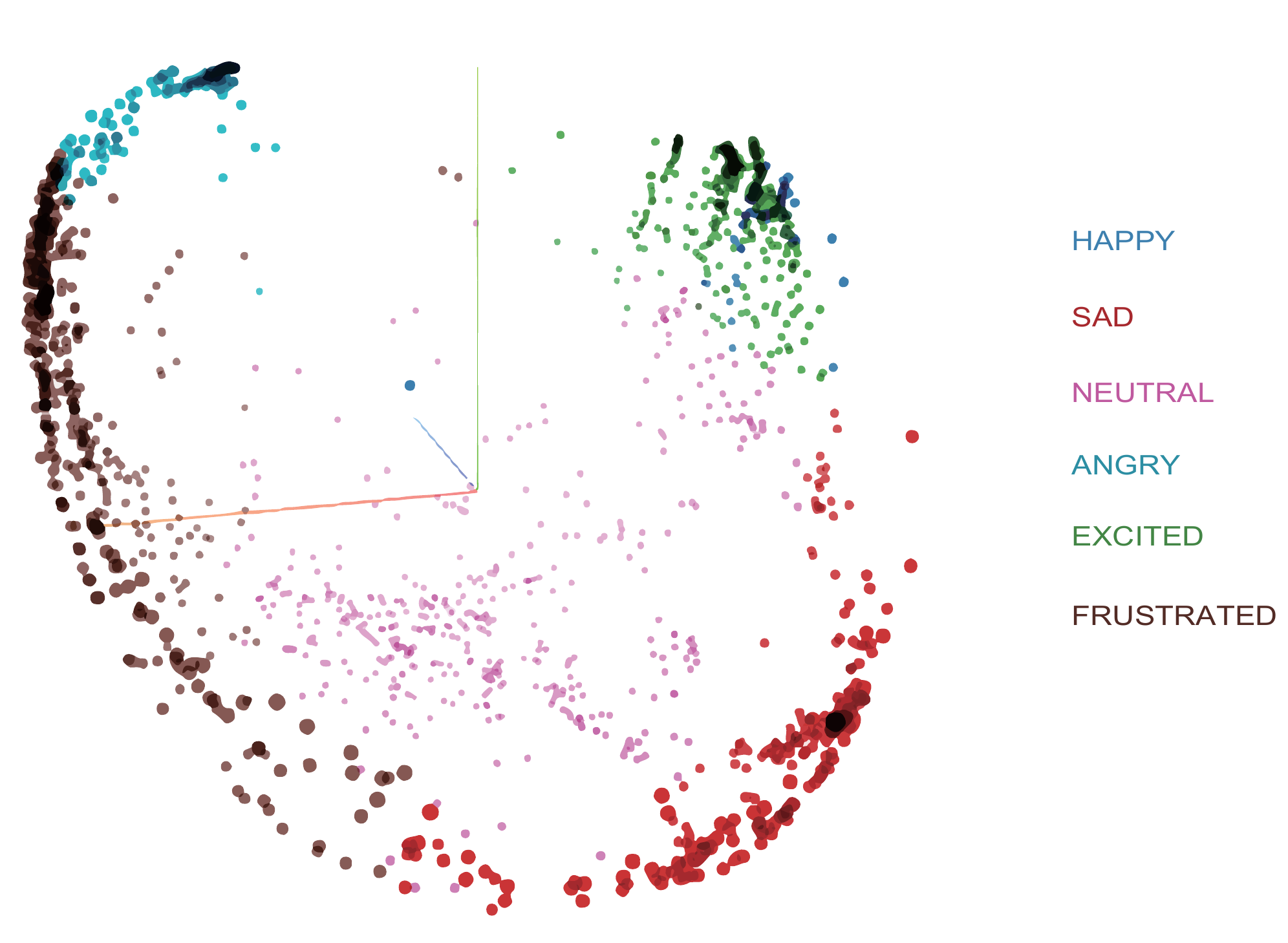}
    \subcaption{DialogueRNN}
\label{fig:Dimensionality Reduction of DialogueRNN}
\end{subfigure}
\linespread{1}
\caption{Visualization of learned emotion features via dimensionality reduction.}
\label{fig:DR}
\end{figure}

Secondly, we perform a deeper analysis of our proposed model and DialogueRNN by visualizing the learned emotion feature representations on IEMOCAP as shown in Fig.~\ref{fig:Dimensionality Reduction of BiERU} and Fig.~\ref{fig:Dimensionality Reduction of DialogueRNN}. Vectors fed into the last dense layer followed by softmax for classification are regarded as emotion feature representations of utterances. We use principal component analysis~\cite{wold1987principal} to reduce the dimension of emotion representations from our model (BiERU-lc) and DialogueRNN. The emotion representation is reduced to be 3-dimensional. In Fig.~\ref{fig:Dimensionality Reduction of BiERU} and Fig.~\ref{fig:Dimensionality Reduction of DialogueRNN}, each color represents a predicted sentiment label and the same color means the same sentiment label. The figures show that our model outperforms on extracting emotion features of utterances labeled "happy", which is consistent with the results in Table 2. In detail, neutral is an intermediate emotion and every other emotion can smoothly transfer into neutral and vice versa. Therefore, in both our BiERU model and DialogueRNN model, neutral has more overlapping regions compared with other emotions. Compared with DialogueRNN, our model distinguishes happy \& excited, frustrated \& angry more clearly. Therefore, our model has the ability to learn better emotion features to some extent.

{}

\subsection{Efficiency~Analysis}
\begin{figure}[ht!]
\centering
\begin{subfigure}{0.45\textwidth}
    \includegraphics[width=\textwidth]{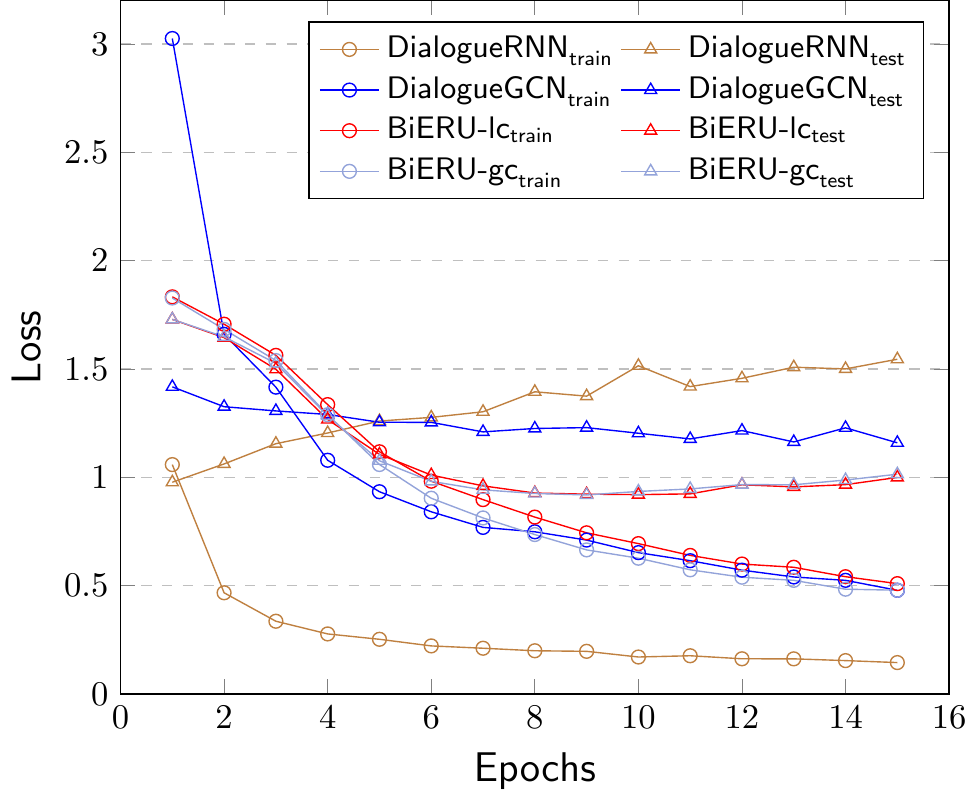}
    \caption{Training curve}
    \label{fig:loss}
\end{subfigure}{}
\begin{subfigure}{0.45\textwidth}
    \includegraphics[width=\textwidth]{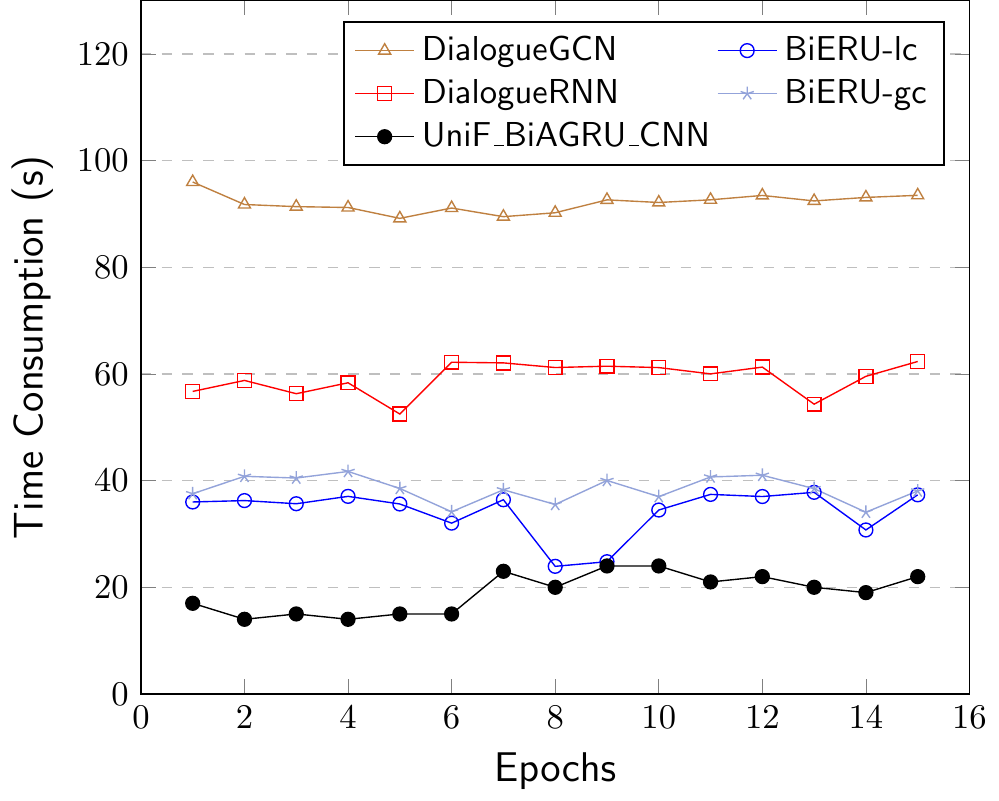}
    \caption{Time consumption}
    \label{fig:time}
\end{subfigure}
\linespread{1}
\caption{Training curve and time consumption logged on a single GPU using the IEMOCAP dataset.}
\end{figure}

We analyze the efficiency of our proposed BiERU model by comparing it with two recent strong baselines. 
Two variants of our model, i.e., BiERU-gc and BiERU-lc, are included.
We choose DialogueRNN and DialogueGCN for comparison as these two are recent competitive methods with public source code. 
Our proposed model has advantages over DialogueRNN, in terms of convergence capacity, the number of trainable parameters, and training time.
In the comparison with DialogueGCN, our models take much less training time. 
Figure~\ref{fig:loss} shows the training curve with training and testing loss plotted. We utilize the same loss function for all the compared models.
Our BiERU-lc and BiERU-gc show comparable convergence speed with their counterparts, while DialogueRNN is prone to overfitting. 

{Our BiERU-gc has fewer trainable parameters and takes less training time than DialogueRNN and DialogueGCN.}
Moreover, BiERU-lc with low-rank matrix approximation has further reduced trainable parameters.
For 100D feature input in the IEMOCAP dataset, our model has about 0.5M parameters, while DialogueRNN requires around 1M. 
For the 600D MELD dataset, DialogueRNN has 2.9M parameters, and our BiERU-lc only has 0.6M. 
With much fewer parameters, our model consequently trains faster than its counterpart as shown in Fig.~\ref{fig:time}, where training time is logged in a single NVIDIA Quadro M5000. 
Our BiERU model with either global or local context is more parameter-efficient and less time-consuming for training.

{}

\subsection{Ablation~Study}
To further explore our proposed BiERU model, we perform an ablation study on its two main components, i.e., GNTB and TFE. We conduct experiments on the IEMOCAP dataset with individual GNTB and TFE modules separately, and their combination, i.e., the complete BiERU. Experimental results on the IEMOCAP dataset are illustrated in Table~\ref{tab:ablation}.

The performance of sole GNTB or TFE is low in terms of accuracy and f1-score. The reason is that outputs of GNTB mainly contain context information and outputs of TFE lack context information. However, when these two modules are combined together as the BiERU model, the accuracy and f1-score increase dramatically, which proves the effectiveness of our BiERU model. More importantly, the GNTB and TFE modules couple significantly well to enhance the performance.
\begin{table}[!h]
\small
    \centering
    \begin{tabular}{|c|c|c|c|}
    \toprule
    GNTB & TFE & ACCURACY & F1-SCORE \\
    \midrule
     - & + & 55.45 &  55.17 \\
     + & - & 49.85 & 49.42\\
     + & + & 65.93 & 64.63 \\
    \bottomrule
    \end{tabular}
    \linespread{1}
    \caption{Results of ablated BiERU on the IEMOCAP dataset. Accuracy and F1-score are weighted average.}
    \label{tab:ablation}
\end{table}{}

\section{Conclusion}
\label{sec:conclusion}
In this paper, we proposed a fast, compact and parameter-efficient party-ignorant framework BiERU for sentiment analysis in conversations. Our proposed GNTB, skilled at context compositionality, reduced the number of parameters and was suitable for different structures. Additionally, our TFE is capable of extracting high-quality emotion features for sentiment analysis. We proved that it is feasible to both simplify the model structure and improve performance simultaneously.

Our model outperforms current state-of-the-art models on three standard datasets in most cases. In addition, our method has the ability to model conversations with arbitrary turns and speakers, which we plan to study further in the future. Finally, we also plan to adopt more recent emotion categorization models, e.g., the Hourglass of Emotions, to better distinguish between similar yet different emotions.

\section*{Acknowledgements}
This research is supported by the Agency for Science, Technology and Research (A*STAR) under its AME Programmatic Funding Scheme (Project \#A18A2b0046).



\begin{thebibliography}{10}
\providecommand{\url}[1]{#1}
\csname url@samestyle\endcsname
\providecommand{\newblock}{\relax}
\providecommand{\bibinfo}[2]{#2}
\providecommand{\BIBentrySTDinterwordspacing}{\spaceskip=0pt\relax}
\providecommand{\BIBentryALTinterwordstretchfactor}{4}
\providecommand{\BIBentryALTinterwordspacing}{\spaceskip=\fontdimen2\font plus
\BIBentryALTinterwordstretchfactor\fontdimen3\font minus
  \fontdimen4\font\relax}
\providecommand{\BIBforeignlanguage}[2]{{%
\expandafter\ifx\csname l@#1\endcsname\relax
\typeout{** WARNING: IEEEtran.bst: No hyphenation pattern has been}%
\typeout{** loaded for the language `#1'. Using the pattern for}%
\typeout{** the default language instead.}%
\else
\language=\csname l@#1\endcsname
\fi
#2}}
\providecommand{\BIBdecl}{\relax}
\BIBdecl

\bibitem{maasur}
Y.~Ma, K.~L. Nguyen, F.~Xing, and E.~Cambria, ``A survey on empathetic dialogue
  systems,'' \emph{Information Fusion}, vol.~64, pp. 50--70, 2020.

\bibitem{majumder2019dialoguernn}
N.~Majumder, S.~Poria, D.~Hazarika, R.~Mihalcea, A.~Gelbukh, and E.~Cambria,
  ``Dialogue{RNN}: An attentive rnn for emotion detection in conversations,''
  in \emph{Proceedings of the AAAI Conference on Artificial Intelligence},
  vol.~33, 2019, pp. 6818--6825.

\bibitem{jiao2019real}
W.~Jiao, M.~Lyu, and I.~King, ``Real-time emotion recognition via attention
  gated hierarchical memory network,'' \emph{Proceedings of the AAAI Conference
  on Artificial Intelligence}, vol.~34, no.~05, pp. 8002--8009, Apr. 2020.

\bibitem{poria2017context}
S.~Poria, E.~Cambria, D.~Hazarika, N.~Majumder, A.~Zadeh, and L.-P. Morency,
  ``Context-dependent sentiment analysis in user-generated videos,'' in
  \emph{Proceedings of the 55th Annual Meeting of the Association for
  Computational Linguistics (Volume 1: Long Papers)}, 2017, pp. 873--883.

\bibitem{hazarika2018conversational}
D.~Hazarika, S.~Poria, A.~Zadeh, E.~Cambria, L.-P. Morency, and R.~Zimmermann,
  ``Conversational memory network for emotion recognition in dyadic dialogue
  videos,'' in \emph{Proceedings of the 2018 Conference of the North American
  Chapter of the Association for Computational Linguistics: Human Language
  Technologies, Volume 1 (Long Papers)}, 2018, pp. 2122--2132.

\bibitem{ghosal2019dialoguegcn}
D.~Ghosal, N.~Majumder, S.~Poria, N.~Chhaya, and A.~Gelbukh, ``{D}ialogue{GCN}:
  A graph convolutional neural network for emotion recognition in
  conversation,'' in \emph{Proceedings of the 2019 Conference on Empirical
  Methods in Natural Language Processing and the 9th International Joint
  Conference on Natural Language Processing (EMNLP-IJCNLP)}.\hskip 1em plus
  0.5em minus 0.4em\relax Hong Kong, China: Association for Computational
  Linguistics, Nov. 2019, pp. 154--164.

\bibitem{mitchell2010composition}
J.~Mitchell and M.~Lapata, ``Composition in distributional models of
  semantics,'' \emph{Cognitive science}, vol.~34, no.~8, pp. 1388--1429, 2010.

\bibitem{partee1995lexical}
B.~Partee, ``Lexical semantics and compositionality,'' \emph{An invitation to
  cognitive science: Language}, vol.~1, pp. 311--360, 1995.

\bibitem{hochreiter1997long}
S.~Hochreiter and J.~Schmidhuber, ``Long short-term memory,'' \emph{Neural
  computation}, vol.~9, no.~8, pp. 1735--1780, 1997.

\bibitem{kim2014convolutional}
Y.~Kim, ``Convolutional neural networks for sentence classification,''
  \emph{arXiv preprint arXiv:1408.5882}, 2014.

\bibitem{cambig}
E.~Cambria, H.~Wang, and B.~White, ``Guest editorial: Big social data
  analysis,'' \emph{Knowledge-Based Systems}, vol.~69, pp. 1--2, 2014.

\bibitem{zhu2020sentivec}
L.~Zhu, W.~Li, Y.~Shi, and K.~Guo, ``Sentivec: Learning sentiment-context
  vector via kernel optimization function for sentiment analysis,'' \emph{IEEE
  Transactions on Neural Networks and Learning Systems}, vol.~32, no.~6, pp.
  2561--2572, 2020.

\bibitem{zadmul}
A.~Zadeh, R.~Zellers, E.~Pincus, and L.-P. Morency, ``Multimodal sentiment
  intensity analysis in videos: Facial gestures and verbal messages,''
  \emph{IEEE Intelligent Systems}, vol.~31, no.~6, pp. 82--88, 2016.

\bibitem{ragima}
E.~Ragusa, C.~Gianoglio, R.~Zunino, and P.~Gastaldo, ``Image polarity detection
  on resource-constrained devices,'' \emph{{IEEE} Intelligent Systems},
  vol.~35, no.~6, pp. 50--57, 2020.

\bibitem{esucro}
A.~Esuli, A.~Moreo, and F.~Sebastiani, ``Cross-lingual sentiment
  quantification,'' \emph{{IEEE} Intelligent Systems}, vol.~35, no.~3, pp.
  106--114, 2020.

\bibitem{weiasp}
A.~Weichselbraun, S.~Gindl, F.~Fischer, S.~Vakulenko, and A.~Scharl,
  ``Aspect-based extraction and analysis of affective knowledge from social
  media streams,'' \emph{IEEE Intelligent Systems}, vol.~32, no.~3, pp. 80--88,
  2017.

\bibitem{banlex}
A.~Bandhakavi, N.~Wiratunga, S.~Massie, and P.~Deepak, ``Lexicon generation for
  emotion analysis from text,'' \emph{IEEE Intelligent Systems}, vol.~32,
  no.~1, pp. 102--108, 2017.

\bibitem{xuuins}
F.~Xu, J.~Yu, and R.~Xia, ``Instance-based domain adaptation via
  multi-clustering logistic approximation,'' \emph{IEEE Intelligent Systems},
  vol.~33, no.~1, pp. 78--88, 2018.

\bibitem{akhnoo}
M.~S. Akhtar, A.~Ekbal, S.~Narayan, and V.~Singh, ``No, that never happened!!
  investigating rumors on twitter,'' \emph{IEEE Intelligent Systems}, vol.~33,
  no.~5, pp. 8--15, 2018.

\bibitem{reisup}
J.~Reis, A.~Correia, F.~Murai, A.~Veloso, and F.~Benevenuto, ``Supervised
  learning for fake news detection,'' \emph{IEEE Intelligent Systems}, vol.~34,
  no.~2, pp. 76--81, 2019.

\bibitem{mihwha}
R.~Mihalcea and A.~Garimella, ``What men say, what women hear: Finding
  gender-specific meaning shades,'' \emph{IEEE Intelligent Systems}, vol.~31,
  no.~4, pp. 62--67, 2016.

\bibitem{buktyp}
A.~Bukeer, G.~Roffo, and A.~Vinciarelli, ``Type like a man! inferring gender
  from keystroke dynamics in live-chats,'' \emph{IEEE Intelligent Systems},
  vol.~34, no.~6, 2019.

\bibitem{yanseg}
Q.~Yang, Y.~Rao, H.~Xie, J.~Wang, F.~L. Wang, and W.~H. Chan, ``Segment-level
  joint topic-sentiment model for online review analysis,'' \emph{IEEE
  Intelligent Systems}, vol.~34, no.~1, pp. 43--50, 2019.

\bibitem{mahdet}
D.~Mahata, J.~Friedrichs, R.~R. Shah, and J.~Jiang, ``Detecting personal intake
  of medicine from twitter,'' \emph{IEEE Intelligent Systems}, vol.~33, no.~4,
  pp. 87--95, 2018.

\bibitem{qurmul}
S.~A. Qureshi, S.~Saha, M.~Hasanuzzaman, and G.~Dias, ``Multitask
  representation learning for multimodal estimation of depression level,''
  \emph{IEEE Intelligent Systems}, vol.~34, no.~5, pp. 45--52, 2019.

\bibitem{ebrcha}
M.~Ebrahimi, A.~Hossein, and A.~Sheth, ``Challenges of sentiment analysis for
  dynamic events,'' \emph{IEEE Intelligent Systems}, vol.~32, no.~5, pp.
  70--75, 2017.

\bibitem{valsen}
A.~Valdivia, V.~Luzon, and F.~Herrera, ``Sentiment analysis in tripadvisor,''
  \emph{IEEE Intelligent Systems}, vol.~32, no.~4, pp. 72--77, 2017.

\bibitem{biicro}
J.-W. Bi, Y.~Liu, and Z.-P. Fan, ``Crowd intelligence: Conducting asymmetric
  impact-performance analysis based on online reviews.'' \emph{{IEEE}
  Intelligent Systems}, vol.~35, no.~2, pp. 92--98, 2020.

\bibitem{duucom}
J.~Du, L.~Gui, R.~Xu, Y.~Xia, and X.~Wang, ``Commonsense knowledge enhanced
  memory network for stance classification,'' \emph{{IEEE} Intelligent
  Systems}, vol.~35, no.~4, pp. 102--109, 2020.

\bibitem{wellea}
C.~Welch, V.~Perez-Rosas, J.~Kummerfeld, and R.~Mihalcea, ``Learning from
  personal longitudinal dialog data,'' \emph{IEEE Intelligent Systems},
  vol.~34, no.~4, pp. 16--23, 2019.

\bibitem{schint}
J.~Schuurmans and F.~Frasincar, ``Intent classification for dialogue
  utterances,'' \emph{{IEEE} Intelligent Systems}, vol.~35, no.~1, pp. 82--88,
  2020.

\bibitem{zhang2020knowledge}
B.~Zhang, X.~Li, X.~Xu, K.-C. Leung, Z.~Chen, and Y.~Ye, ``Knowledge guided
  capsule attention network for aspect-based sentiment analysis,''
  \emph{IEEE/ACM Transactions on Audio, Speech, and Language Processing},
  vol.~28, pp. 2538--2551, 2020.

\bibitem{liu2021speech}
D.~Liu, L.~Chen, Z.~Wang, and G.~Diao, ``Speech expression multimodal emotion
  recognition based on deep belief network,'' \emph{Journal of Grid Computing},
  vol.~19, no.~2, pp. 1--13, 2021.

\bibitem{camnt6}
E.~Cambria, Y.~Li, F.~Xing, S.~Poria, and K.~Kwok, ``{SenticNet} 6: Ensemble
  application of symbolic and subsymbolic {AI} for sentiment analysis,'' in
  \emph{{CIKM}}, 2020, pp. 105--114.

\bibitem{ragheb2019attention}
W.~Ragheb, J.~Az{\'e}, S.~Bringay, and M.~Servajean, ``Attention-based modeling
  for emotion detection and classification in textual conversations,''
  \emph{arXiv preprint arXiv:1906.07020}, 2019.

\bibitem{sukhbaatar2015end}
S.~Sukhbaatar, J.~Weston, R.~Fergus \emph{et~al.}, ``End-to-end memory
  networks,'' in \emph{Advances in neural information processing systems},
  2015, pp. 2440--2448.

\bibitem{hazarika2018icon}
D.~Hazarika, S.~Poria, R.~Mihalcea, E.~Cambria, and R.~Zimmermann, ``{ICON}:
  Interactive conversational memory network for multimodal emotion detection,''
  in \emph{Proceedings of the 2018 Conference on Empirical Methods in Natural
  Language Processing}, 2018, pp. 2594--2604.

\bibitem{ghosal2020cosmic}
D.~Ghosal, N.~Majumder, A.~Gelbukh, R.~Mihalcea, and S.~Poria, ``{COSMIC}:
  {CO}mmon{S}ense knowledge for e{M}otion identification in conversations,'' in
  \emph{Findings of the Association for Computational Linguistics: EMNLP
  2020}.\hskip 1em plus 0.5em minus 0.4em\relax Online: Association for
  Computational Linguistics, Nov. 2020, pp. 2470--2481.

\bibitem{wang2019capturing}
S.~Wang, G.~Peng, Z.~Zheng, and Z.~Xu, ``Capturing emotion distribution for
  multimedia emotion tagging,'' \emph{IEEE Transactions on Affective
  Computing}, no.~01, pp. 1--1, feb 5555.

\bibitem{zhong2019knowledge}
P.~Zhong, D.~Wang, and C.~Miao, ``Knowledge-enriched transformer for emotion
  detection in textual conversations,'' in \emph{Proceedings of the 2019
  Conference on Empirical Methods in Natural Language Processing and the 9th
  International Joint Conference on Natural Language Processing
  (EMNLP-IJCNLP)}, 2019, pp. 165--176.

\bibitem{qin2020dcr}
L.~Qin, W.~Che, Y.~Li, M.~Ni, and T.~Liu, ``Dcr-net: A deep co-interactive
  relation network for joint dialog act recognition and sentiment
  classification.'' in \emph{AAAI}, 2020, pp. 8665--8672.

\bibitem{socher2013reasoning}
R.~Socher, D.~Chen, C.~D. Manning, and A.~Ng, ``Reasoning with neural tensor
  networks for knowledge base completion,'' in \emph{Advances in neural
  information processing systems}, 2013, pp. 926--934.

\bibitem{socher2013recursive}
R.~Socher, A.~Perelygin, J.~Wu, J.~Chuang, C.~D. Manning, A.~Ng, and C.~Potts,
  ``Recursive deep models for semantic compositionality over a sentiment
  treebank,'' in \emph{Proceedings of the 2013 conference on empirical methods
  in natural language processing}, 2013, pp. 1631--1642.

\bibitem{busso2008iemocap}
C.~Busso, M.~Bulut, C.-C. Lee, A.~Kazemzadeh, E.~Mower, S.~Kim, J.~N. Chang,
  S.~Lee, and S.~S. Narayanan, ``{IEMOCAP}: Interactive emotional dyadic motion
  capture database,'' \emph{Language resources and evaluation}, vol.~42, no.~4,
  p. 335, 2008.

\bibitem{nair2010rectified}
V.~Nair and G.~E. Hinton, ``Rectified linear units improve restricted boltzmann
  machines,'' in \emph{Proceedings of the 27th international conference on
  machine learning (ICML-10)}, 2010, pp. 807--814.

\bibitem{graves2005framewise}
A.~Graves and J.~Schmidhuber, ``Framewise phoneme classification with
  bidirectional lstm and other neural network architectures,'' \emph{Neural
  networks}, vol.~18, no. 5-6, pp. 602--610, 2005.

\bibitem{li2020user}
W.~Li, L.~Zhu, Y.~Shi, K.~Guo, and E.~Cambria, ``User reviews: Sentiment
  analysis using lexicon integrated two-channel cnn-lstm family models,''
  \emph{Applied Soft Computing}, vol.~94, p. 106435, 2020.

\bibitem{kingma2014adam}
D.~P. Kingma and J.~Ba, ``Adam: A method for stochastic optimization,''
  \emph{arXiv preprint arXiv:1412.6980}, 2014.

\bibitem{schuster1997bidirectional}
M.~Schuster and K.~K. Paliwal, ``Bidirectional recurrent neural networks,''
  \emph{IEEE transactions on Signal Processing}, vol.~45, no.~11, pp.
  2673--2681, 1997.

\bibitem{schuller2012avec}
B.~Schuller, M.~Valster, F.~Eyben, R.~Cowie, and M.~Pantic, ``{AVEC} 2012: the
  continuous audio/visual emotion challenge,'' in \emph{Proceedings of the 14th
  ACM international conference on Multimodal interaction}.\hskip 1em plus 0.5em
  minus 0.4em\relax ACM, 2012, pp. 449--456.

\bibitem{poria2018meld}
S.~Poria, D.~Hazarika, N.~Majumder, G.~Naik, E.~Cambria, and R.~Mihalcea,
  ``{MELD: A} multimodal multi-party dataset for emotion recognition in
  conversations,'' in \emph{{ACL}}, 2019, pp. 527--536.

\bibitem{susanto2020hourglass}
Y.~Susanto, A.~G. Livingstone, B.~C. Ng, and E.~Cambria, ``The hourglass model
  revisited,'' \emph{IEEE Intelligent Systems}, vol.~35, no.~5, pp. 96--102,
  2020.

\bibitem{mckeown2011semaine}
G.~McKeown, M.~Valstar, R.~Cowie, M.~Pantic, and M.~Schroder, ``The semaine
  database: Annotated multimodal records of emotionally colored conversations
  between a person and a limited agent,'' \emph{IEEE Transactions on Affective
  Computing}, vol.~3, no.~1, pp. 5--17, 2011.

\bibitem{cho2014learning}
K.~Cho, B.~van Merrienboer, C.~Gulcehre, D.~Bahdanau, F.~Bougares, H.~Schwenk,
  and Y.~Bengio, ``Learning phrase representations using rnn encoder--decoder
  for statistical machine translation,'' in \emph{Proceedings of the 2014
  Conference on Empirical Methods in Natural Language Processing (EMNLP)},
  2014, pp. 1724--1734.

\bibitem{srivastava2014dropout}
N.~Srivastava, G.~Hinton, A.~Krizhevsky, I.~Sutskever, and R.~Salakhutdinov,
  ``Dropout: a simple way to prevent neural networks from overfitting,''
  \emph{The journal of machine learning research}, vol.~15, no.~1, pp.
  1929--1958, 2014.

\bibitem{paszke2019pytorch}
A.~Paszke, S.~Gross, F.~Massa, A.~Lerer, J.~Bradbury, G.~Chanan, T.~Killeen,
  Z.~Lin, N.~Gimelshein, L.~Antiga \emph{et~al.}, ``Pytorch: An imperative
  style, high-performance deep learning library,'' in \emph{Advances in Neural
  Information Processing Systems}, 2019, pp. 8024--8035.

\bibitem{wold1987principal}
S.~Wold, K.~Esbensen, and P.~Geladi, ``Principal component analysis,''
  \emph{Chemometrics and intelligent laboratory systems}, vol.~2, no. 1-3, pp.
  37--52, 1987.

\end{thebibliography}

\end{document}